\documentclass[runningheads]{llncs}
\usepackage[T1]{fontenc}
\usepackage{amsmath}
\usepackage{graphicx}
\begin{document}
\title{Exploring Various Sequential Learning Methods for Deformation History Modeling\thanks{Supported by organization x.}}
\author{Muhammed Adil YATKIN\inst{1}\orcidID{0000-0001-6776-8813} \and
Mihkel Korgesaar\inst{1}\orcidID{0000-0003-3920-9031} \and
Jani Romanoff\inst{2}\orcidID{2222--3333-4444-5555} \and
Joshua Stuckner\inst{3}\orcidID{0000-0002-4642-0225}\and
Ümit Işlak\inst{4}\orcidID{0000-0003-4281-5171}\and
Hasan Kurban\inst{5}\orcidID{0000-0003-3142-2866}}
\authorrunning{F. Author et al.}
\institute{Tallinn University of Technology
\email{muyatk@taltech.ee}\\
\url{https://taltech.ee/en/kuressaare-college} \and
Aalto University\\
\email{jani.romanoff@aalto.fi} \and
Nasa Research Glenn Center\\
\email{joshua.stuckner@nasa.gov}\and
Bogazici University\\
\email{umit.islak1@bogazici.edu.tr}\and
College of Science and Engineering, Hamad Bin Khalifa University\\
\email{hkurban@hbku.edu.qa}}
\maketitle              % typeset the header of the contribution
\begin{abstract}
Current neural network (NN) models can learn patterns from data points with historical dependence. Specifically, in natural language processing (NLP), sequential learning has transitioned from recurrence-based architectures to transformer-based architectures. However, it is unknown which NN architectures will perform the best on datasets containing deformation history due to mechanical loading. Thus, this study ascertains the appropriateness of 1D-convolutional, recurrent, and transfor\-mer-based architectures for predicting deformation localization based on the earlier states in the form of deformation history. Following this investigation, the crucial incompatibility issues between the mathematical computation of the prediction process in the best-performing NN architectures and the actual values derived from the natural physical properties of the deformation paths are examined in detail.
\end{abstract}

\section{Introduction}
Finite Element (FE) simulations are widely used in engineering to analyze the mechanical behavior of structures under various loading conditions. These simulations enable the replication of real-world physical phenomena, providing a foundation for physics and engineering-driven calculations under specific boundary conditions. One important application is predicting localization during sheet metal forming under bilinear loading paths, which was the focus of our previous work \cite{Yatkin2024}.

Recent advances have demonstrated that certain components of FE simulations can be replaced by machine learning (ML) models acting as surrogates, allowing for the prediction of complex nonlinear relationships while significantly reducing computational cost. Some studies focus on designing customized ML approaches \cite{BonattiMohr-1-for-all2021,Bonatti-consistency2022,yatkin2025topological} tailored to the intrinsic properties of simulation derived data, whereas others rely on conventional methods such as support vector machines (SVM), regression models \cite{samadian2024application}, and well-established neural network architectures \cite{meethal2023finite,sunil2024fepinnsfiniteelementbasedphysicsinformedneural}.

In our previous work \cite{Yatkin2024}, we formulated a mathematical mapping to predict sheet metal localization by relating strain, stress, and damage states. We trained a one-dimensional convolutional neural network (1D CNN) to predict localization points with high accuracy, eliminating the need for additional simulations. While surrogate modeling has been extensively studied, a direct comparison of different sequential learning architectures for deformation history learning remains underexplored. In this study, we address this gap by evaluating the learning capabilities of three well-known architectures 1D CNNs, Transformers, and Recurrent Neural Networks (RNNs) to establish a stronger baseline for future research.

\subsection{Machine Learning Methods}
Unlike the 1D CNN used in our previous study \cite{Yatkin2024}, recent investigations in material modeling have preferred other deep learning-based surrogates for constitutive models to capture deformation history in numerical simulations. Given the sequential nature of the problem, recurrent neural networks (RNNs) have gained attention for learning complex input-output mappings, such as learning stress-strain relationships based on deformation history. For instance, Mozaffar et al. \cite{Mozaffar2019DeepLP} suggested that RNNs simplify plasticity theory formulations, while Wu et al. \cite{Wu2020} demonstrated their effectiveness as surrogates for meso-scale Boundary Value Problems (BVP) within the realm of computational multiscale analysis. Ghavamian and Simone \cite{Ghavamian2019} accelerated multiscale FE simulations of history-dependent materials by replacing a micromechanical model with an RNN surrogate. 

Conventional constitutive models have been replaced or enriched by data-driven methods \cite{Chinesta2017,WANG2024112900,Hartmaier2020,DAAREYNI2022111710}. Tabarraei et al. \cite{ELAPOLU2022110878} used a hybrid NN architecture combining 2D-convolutional (2D CNNs), bidirectional RNNs, and fully connected layers to predict crack growth with high accuracy. Moreover, Alahyarizadeh et al. \cite{PEIVASTE2022111750} developed a U-net-based \cite{DBLP:journals/corr/RonnebergerFB15} NN based surrogate model to reduce the computational cost associated with phase-field simulations for microstructure evolution. However, no studies have systematically compared the accuracy and applicability of the three main sequential learning approaches in neural netwworks: 1D CNNs, RNNs, and Transformers.

This study aims to investigate and compare three prominent structured NN architectures for sequential learning: 1D CNNs, encoder-decoder-based RNNs, and transformers. These architectures are evaluated by training them on bilinear deformation paths generated from numerical simulations and analyzing their convergence behavior. Furthermore, the best-performing NN architectures are examined in detail to assess their predictions' compatibility with the deformation paths' natural physical properties. 

\section{Prior Work}
\begin{figure}[!h]
\centering
\includegraphics[scale=1.0]{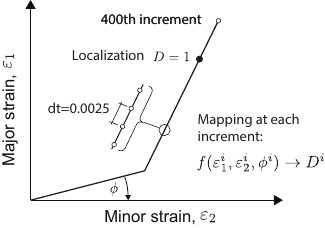}
\caption{The mapping defined in Yatkin and K\~orgesaar \cite{Yatkin2024} from bilinear strain 
paths to the damage indicator D.}
\label{fig:P1figure}
\end{figure}

In our previous study \cite{Yatkin2024}, a bilinear loading dataset is generated with physics-based Marciniak-Kuczynski (MK) FE-based numerical simulations. The dataset contains 19098 bi-linear non-proportional loading paths composed of two proportional segments with known characteristics as shown in Fig. \ref{fig:P1figure}. Each loading path consists of 400 increments recorded at a fixed sampling rate of 0.0025 seconds. The equivalent plastic strain $\bar{\varepsilon}_{i}$ is calculated at each increment with eq. \eqref{eqn:deps} from which damage indicator $D^{i}$ in eq.(\ref{eqn:norm_eps______}) is evaluated. This information was cast into sequential ML problem by mapping the strain state and loading direction $\phi$ at each increment into a damage indicator $D^{i}$ as $f(\varepsilon_1^i, \varepsilon_2^i, \phi^i) \rightarrow D^i$ and was solved with 1D CNN based architecture trained on the deformation history.

\begin{equation}
\label{eqn:deps}
\bar{\varepsilon}_{i}=\int_0^i \mathrm{d} \bar{\varepsilon} \text { where } \mathrm{d} \bar{\varepsilon}=\frac{2}{\sqrt{3}} \sqrt{\left(\mathrm{d} \varepsilon_1\right)^2+\left(\mathrm{d} \varepsilon_{2}\right)^2+\left(\mathrm{d} \varepsilon_1\right)\left(\mathrm{d} \varepsilon_{2}\right)},
\end{equation}

\begin{equation}
D^{i} =\frac{{\bar{\varepsilon}}_i}{\bar{\varepsilon}_{fail}}  \;\;,\;\; i \in {1,2,....,400}.
\label{eqn:norm_eps______}
\end{equation}

While 1D CNN based architectures are effective at capturing local patterns in sequential data points, RNNs and transformers are better at capturing temporal and long-term dependencies, which is more relevant for history learning related to load history in mechanical engineering problems. In addition, finite-element-based numerical simulations rely on incremental processing, similar to recurrent neural networks in handling sequential data. Conversely, 1D CNNs do not process data incrementally. Instead, they simultaneously apply convolutional filters to the entire sequence, capturing local patterns within the filters' receptive fields. The primary limitation of 1D CNNs for incremental computations over time is their simultaneous processing nature and the lack of a mechanism to maintain and update an internal state over time. 

In contrast, RNNs are specifically designed to process sequential data and maintain temporal context, rendering them more suitable for tasks involving time-dependent data. Driven by recent trends in using RNNs to solve history-dependent constitutive behavior (section 1.2) and their efficiency in handling sequential data, we tested different sequential learning architectures to improve the accuracy and efficiency of surrogate models, namely encoder decoder based RNNs, 1D CNNs, and Transformers. These architectures were created and trained on the bilinear loading dataset provided in \cite{Yatkin2024}, and their relative accuracies were compared. This comparative study aims to identify which architectures are better at learning the mapping between the bilinear loading paths in the strain space and the amount of damage.

\section{Neural Network architectures for history dependent damage}

The input sequence length used for the 1D CNN architecture for history learning in the study by Yatkin and Kõrgesaar \cite{Yatkin2024} was fixed. However, in actual practical application cases where surrogates are embedded in numerical FE simulations to process sequential incremental data, the current damage state prediction is based on the previous strain history, which is continuously changing. Therefore, the starting point for this investigation was in the NN architectures that can handle sequence-to-sequence data regardless of the input sequence length.

\subsection{Sequential Learning Approaches}
\label{SequentiaL_Learning}
\begin{figure}[!h]
\centering
\includegraphics[scale=1.0]{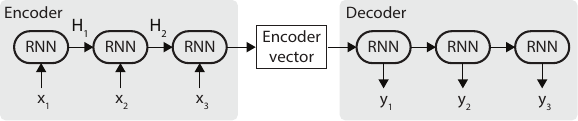}
\caption{While the encoder part generates the final hidden state, the decoder part uses it as an initial hidden state and calculates the outputs.}
\label{fig:Encoder-Decoder-Architecture}
\end{figure}

NN architectures developed for sequential learning problems in non-mechanical applications commonly include the encoder-decoder principle. 
Examples of these are frequent in tasks such as machine translation \cite{Zhu2020}, named entity recognition \cite{Yamada2020}, and image captioning \cite{Li2022}. The advantage of encoder-decoder principle-based architectures is that they allow the processing of variable input and output length sequences. The encoder part encodes the input sequence into a fixed vector, while the decoder part uses this vector to generate the output sequence shown in Fig. \ref{fig:Encoder-Decoder-Architecture}. The encoder-decoder-based architectures consist of layers such as Recurrent, 1D Convolutional and Attention layers. The most suitable and best-performing layer architecture depends on the given sequence-to-sequence problem and must be discovered by testing. 
For instance, in the NLP field the general trend went from RNN-based architectures to Transformers-based architectures because of their higher accuracy (notably chatGPT). Although several studies \cite{BonattiMohr-1-for-all2021,Bonatti-consistency2022} focused more on customized RNN approaches for material engineering-related problems, here, we investigated the learning ability of different NN architectures that are suitable for sequential learning on deformation history.

%To investigate the applicability of the three most common NN structures for sequence-to-sequence modeling, different architectures are built that only include RNN, 1D CNN, and Transformer based NN blocks. These architectures are trained on the previously generated bilinear loading dataset with a difference that the damage indicator for each bilinear path is mapped into $[0,1]$ with an eq. \ref{Convert}. This transformation was necessary because a linear layer with a sigmoid activation is used as the last layer for each built architecture. The function used for this transformation shown in eq. \ref{Convert} is invertible as shown in eq. \ref{Convert_inverse}. Because the predictions obtained from the NN architectures after training are between [0,1] and still need to be converted into damage indicator estimates. 

To investigate the applicability of the three most common NN architectures for sequence-to-sequence modeling, we built three NN architectures, each containing only RNN, 1D CNN, or Transformer layers. These architectures are trained on the previously generated bilinear loading dataset, and their performances are compared by hypertuning each built architecture.

The TensorFlow \cite{tensorflow_developers_2023_8306789} 2.6.0 and Keras 2.15.0 frameworks are used for the implementation process. All trainings were conducted on Nvidia RTX2080 Graphic Processing Unit (GPU) card.

%\begin{equation}
%\label{Convert}
%\begin{aligned}
%f(D^{i}) =& 1 - e^{-D^{i}}
%\end{aligned}
%\end{equation}

%\begin{equation}
%\label{Convert_inverse}
%g(f(D^{i})) = -\log(1 - f(D^{i})) 
%\end{equation}

%\begin{figure}[!h]
%\centering
%\includegraphics[scale=0.8]{Blocks_Representation.pdf}
%\caption{Schematic representation of the NN blocks built from 1D CNN, Recurrent, and Transformer structures. The RNN block shown in (A) composed of two GRU layers. In the 1D CNN block shown in (B), the input is processed through convolutional layers and passes to the GLU layer, which acts as a gating mechanism. When the NN architecture includes more than one block, the output features of the first block are passed to the second block, and so on. Transformer block shown in (C) implemented based on the encoder of the network presented in the study \cite{Vaswani2017}.}
%\label{fig:Encoder-Decoder-Architecture}
%\end{figure}

\subsubsection{Recurrent Neural Networks}
\label{section:RNN}

Recurrent Neural Networks (RNNs) were specifically designed to model sequential data. During the learning process, they encodes learned information into a fixed-size state vector, which is incrementally updated at each time step.. Estimates are computed using this learned state vector through output functions. The computational process of the operating principle of RNNs can be generally formulated as in eq. \eqref{RNN_transition} and \eqref{RNN_output}, where $F$ is the transition function, $H_{t}$ is the state vector, O is the output function, $x_{t}$ is the input value for each time step, and $y_{t}$ is the output at each output time step \cite{a81cdc63b81a42f5af92c81179c94532}.

\begin{equation}
\label{RNN_transition}
H_{t} = F(H_{t-1}, x_{t})
\end{equation}
\begin{equation}
\label{RNN_output}
y_{t} = O(H_{t})
\end{equation}

RNNs can be distinguished by the different transition functions in eq. \ref{RNN_transition} that they use. The most commonly used RNNs in non-mechanical applications are the Gated Recurrent Unit (GRU) and the Long Short Term Memory (LSTM), which consists of a "gating mechanism" in the transition function. This mechanism essentially determines how much information is transformed into the new state vector in each incremental update. While the GRU contains two gates, the update and reset, the LSTM includes three gates, which are the input, output, and forget gates, resulting in the LSTM having a greater number of parameters. Typically, GRU and LSTM-based NN architectures give similar results for sequential learning tasks. For this reason, we preferred to use the GRU layer due to its fewer parameters, making its usage more efficient.

\begin{figure}[!h]
\centering
\includegraphics[scale=1.0]{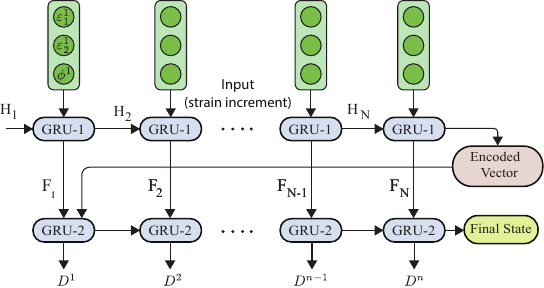}
\caption{Two layered recurrent neural network (RNN) architecture}
\label{fig:Encoder-Decoder-RNN-Architecture}
\end{figure}

\begin{figure}[!h]
\centering
\includegraphics[scale=0.6]{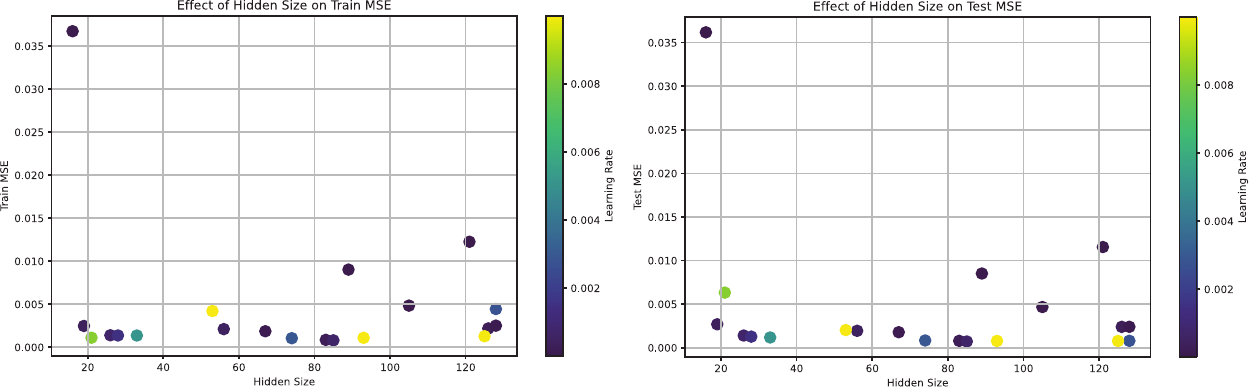}
\caption{RNN architecture training results on the bilinear loading paths}
\label{fig:RNN_Results}
\end{figure}

Here, we formed an NN architecture, shown in Fig. \ref{fig:Encoder-Decoder-RNN-Architecture}, consisting of two GRU layers. The first layer encodes the learned information from the sequential input into the state vector, and the second layer begins the learning process with this encoded vector, continuing the learning on sequential features. The architecture is trained using the Adam optimizer, with hyperparameters tuned through Bayesian optimization, including a learning rate ranging from 0.00001 to 0.001, hidden units varying between 16 and 128, and batch size between 16 and 256. We used an early stopping criterion with a patience of 5 and trained the architecture for 100 epochs. The detailed results based on the bayesian optimization with changing hyperparameters can be seen from Fig. \ref{fig:RNN_Results}.

\begin{figure}[!h]
\centering
\includegraphics[scale=0.8]{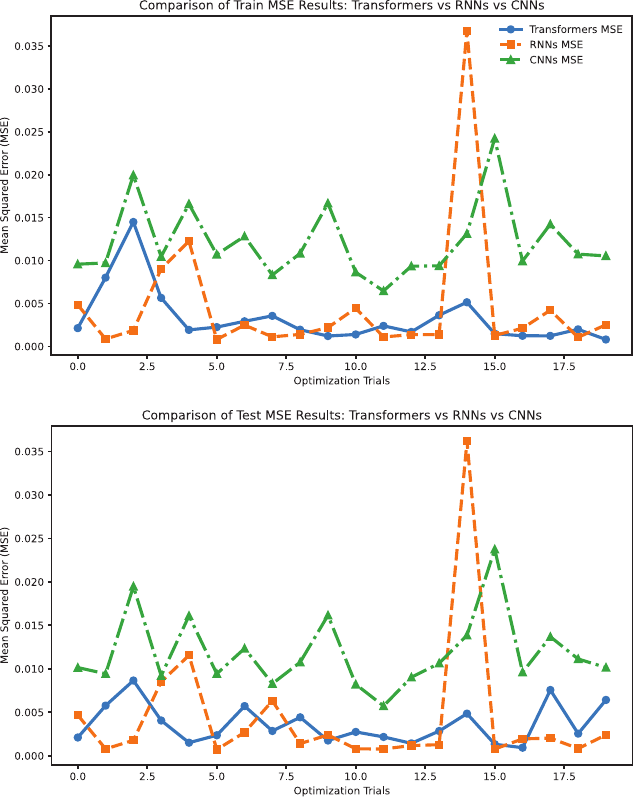}
\caption{Comparison results between 1D-CNN, RNNs, and Transformers based NN architectures on the learning of bilinear deformation history paths.}
\label{fig:Architectural_Comparison}
\end{figure}

As observed from the training MSE results shown in Fig. \ref{fig:RNN_Results}, as the hidden size increases from 20 to 80, the MSE starts to flatten out, and there is no notable improvement beyond 80. Therefore, the most optimal results are obtained with a hidden size between 60 and 80. A similar trend can be observed in the test set MSE results. Additionally, the architecture demonstrates stable convergence in each Bayesian optimization trial shown in Fig. \ref{fig:Architectural_Comparison}, except for a noticeable fluctuation in the 14th trial. However, overall, the architecture exhibited stable convergence across trials, proving that the Encoder-Decoder-based GRU architecture effectively learns bilinear deformation paths.

\subsubsection{Convolutional Networks}
\label{section: 1DNN}
Several studies \cite{10.5555/3305381.3305510,KIRANYAZ2021107398} have shown that NN architectures based on 1D Convolutional layers (1D CNNs) can serve as an efficient alternative  solution for sequential learning tasks; in particular, they can reduce the complexity of encoder-decoder-based RNN architectures and provide a compact solution.

\begin{figure}[!h]
\centering
\includegraphics[scale=0.6]{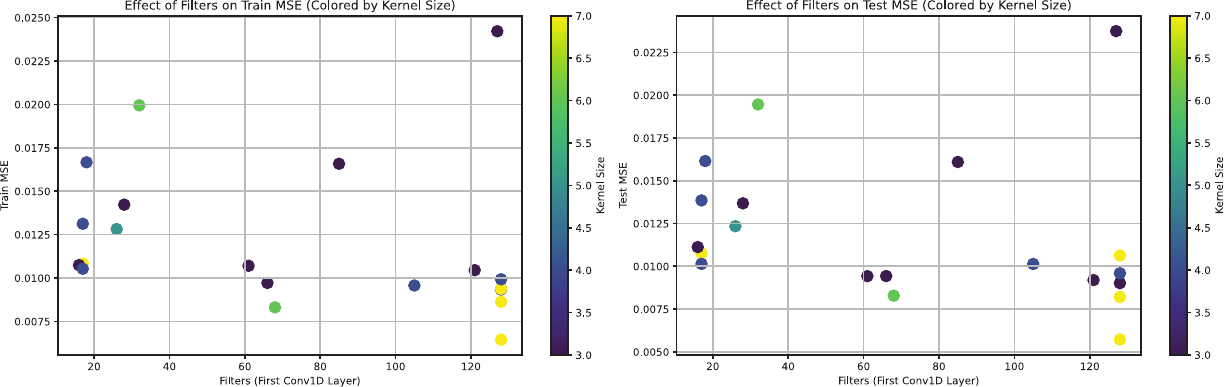}
\caption{1D-CNN Training results on the bilinear loading paths}
\label{fig:1DCNN_Results}
\end{figure}

Here, we have designed an NN architecture utilizing 1D convolutional layers capable of processing arbitrary input data of different lengths. The architecture includes two 1D CNN layers, where the second layer uses a filter of 1. We varied the number of filters between 16 and 128, the kernel size between 3 and 7, the learning rate between 0.00001 and 0.001, and the batch size between 16 and 256, using a Bayesian optimization algorithm. The architecture was trained for
100 epochs with early stopping, using a patience of 5. The detailed results can be observed from Fig. \ref{fig:Architectural_Comparison}, and \ref{fig:1DCNN_Results}. As can be seen from the change in the loss during optimization trials, it fluctuates a lot, and didnt show well convergence, opposite the results of Encoder-Decoder based RNNs.

\subsubsection{Transformers}
\label{section:Transformers}

Transformers were originally introduced in \cite{Vaswani2017} and have shown remarkable performance in sequential learning of real-world problems in a field such as computer vision \cite{DBLP:journals/corr/abs-2101-01169}, natural language processing \cite{DBLP:journals/corr/abs-1810-04805}, etc. Transformers-based architectures rely on attention layers that are not sequence-length dependent on the number of operations like recurrent layers, giving them a training time advantage.

\begin{figure}[!h]
\centering
\includegraphics[scale=1.0]{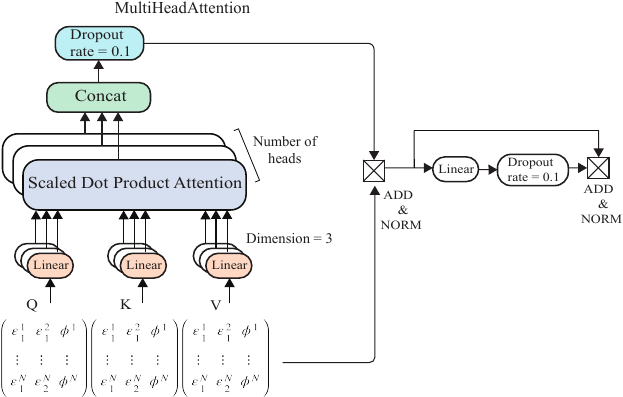}
\caption{Transformers block, which is formed from the encoder state from the paper \cite{Vaswani2017}.}
\label{fig:Transformers_Architecture}
\end{figure}

Here, we constructed a neural network architecture based on multi-head attention and a linear layer, as shown in Fig. \ref{fig:Transformers_Architecture}, which corresponds to the encoder part of the transformer architecture from \cite{Vaswani2017}. The query (Q), key (K), and value (V) are taken from the input sequence. These are projected through linear layers with a dimension of three and then applied to a Scaled Dot-Product Attention, as defined by the number of heads in eq. \eqref{Scaled-Dot-Product-Attention}. After this, the outputs are concatenated, and some neurons are dropped out with a probability rate of 0.1, similar to the approach in \cite{Vaswani2017}. The original input sequence and the obtained features are added together and normalized. The final output is passed through the linear layer, with dropout again applied at a rate of 0.1. In the last part of the block, the output features from both the first and second dropout layers are added and normalized \cite{Kaiming2015}. We adjusted the learning rate between 0.00001 and 0.01, the embedding dimension from 16 to 128, the number of heads from 1 to 8, the feedforward layers from 32 to 256, and the batch size from 16 to 256 through Bayesian optimization during training. The architecture was trained for 100 epochs with early stopping, using a patience of 5.

\begin{equation}
\label{Scaled-Dot-Product-Attention}
Attention(Q, K, V) = softmax(\frac{QK^{T}}{\sqrt{d_{k}}})V
\end{equation}

\begin{figure}[!h]
\centering
\includegraphics[scale=0.6]{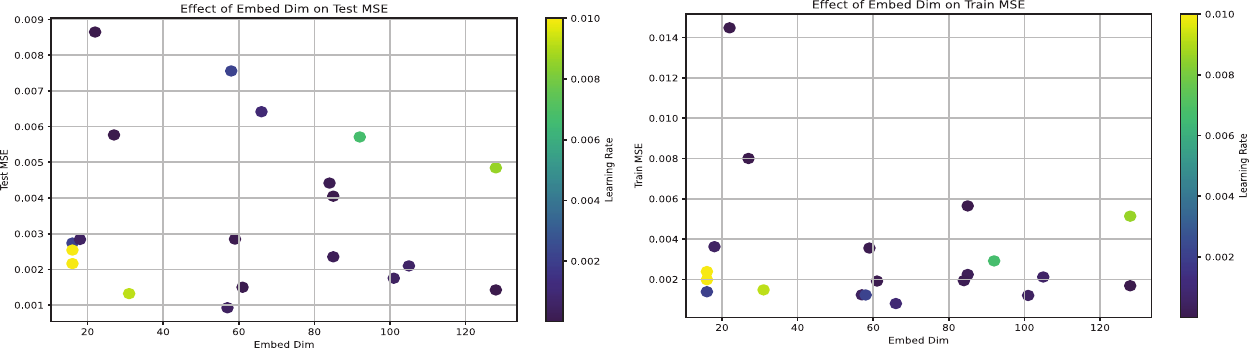}
\caption{Results from the Transformers based neural network ( NN ) architecture.}
\label{fig:Transformers_Results}
\end{figure}

As observed from the detailed results shown in Fig. \ref{fig:Transformers_Results}, the transformer-based architecture demonstrated strong convergence throughout the optimization trials, reducing the loss value across iterations.Additionally, the results for varying embedding dimensions on both the train and test sets indicate that the most optimal performance, with the lowest MSE values, is achieved with a hidden dimension size between 100 and 120. Furthermore, the close alignment of train and test MSE scores confirms the absence of overfitting. Overall, the results highlight that the transformer-based architecture effectively learns bilinear loading paths.

%Limiting the number of parameters to a maximum of 10000, eighty-one different NN architectures are created by varying the number of heads by 2 increments between 2 and 80 and the number of blocks in each architecture. The built NN architectures are trained for 100 epochs on the bilinear loading dataset using the Adam optimizer with a learning rate of 0.0001 and MSE loss. The yielded MSE results on the test set are shown in Fig. \ref{fig:mse_rnn_cnn_trans} (A) and (B) and detailed results are demonstrated in Fig. \ref{fig:mse_rnn_cnn_trans} (E) and (H).

%\begin{figure}[!h]
%\centering
%\includegraphics[scale=0.8]{Architectures_Comparison.pdf}
%\caption{Detailed comparison results betwwen 1D-CNN, RNNs, and Transformers based NN architectures on the learning of bilinear deformation history paths.}
%\label{fig:Architectural_Comparison}
%\end{figure}

\subsection{Discussion of the Results}
\label{section:Discussion}

In addition anlyzing the detailed results on the learning capabilities of different RNN, Transformers, and 1D-CNN-based NN architectures, we also compared their MSE results on both the training and test sets across 20 optimization trials, as shown in Fig. \ref{fig:Architectural_Comparison}. The results reveal that the 1D CNN architecture  demonstrates the weakest learning performance, while the Transformer- and RNN-based models demonstrate strong convergence. The Transformer model provides more stable MSE values throughout the trials, whereas the encoder-decoder RNNs generally yield lower MSE values, making them the most suitable neural network architecture for learning deformation history, with a close approximation to the Transformer-based architecture. Moreover, the MSE values for both the training and test sets are closely aligned, indicating no overfitting in the training process, which further supports the validity of the results obtained.

\begin{figure}[!h]
\centering
\includegraphics[scale = 0.7]
{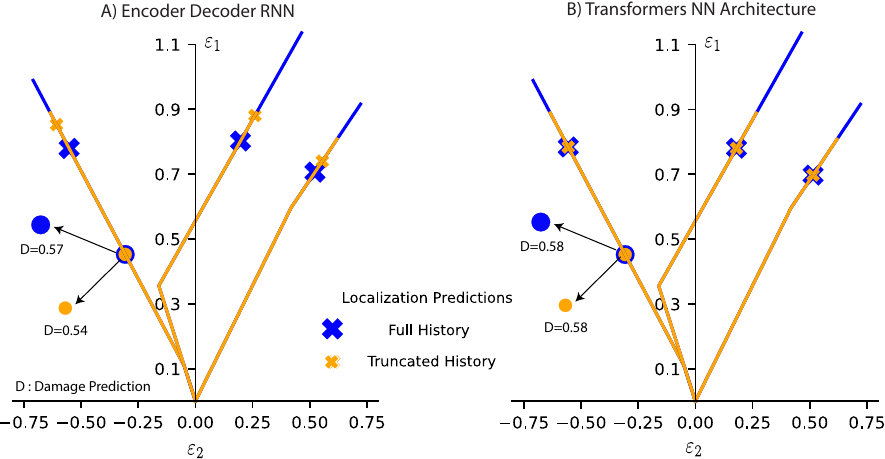}
\caption{Examples of the truncation problem for different bilinear paths shown in A). As can be seen from the plotted results, different localization predictions are obtained by the encoder-decoder based RNN network for the full history and truncated history inputs of the bilinear paths, and for the same time step the damage predictions (D) are different, which is incompatible with the physical state of the deformation history. On the other hand, as shown in B), the localization and damage predictions for the Transformers NN architecture are the same for full history and truncated history inputs.}
\label{fig:Truncation}
\end{figure}

\section{Incompatibility of Encoder-Decoder RNNs with physical state}

In Section \ref{section:Discussion}, we presented the training results of the different architectures. This section focuses on the applicability of encoder-decoder RNN-based NN architectures to real-world use cases. Specifically, when a NN architecture is used to estimate the material damage state during deformation, the estimate must align with the physical state, taking into account the entire (or previous) deformation history. In other words, the current damage estimate produced by the NN architecture should remain consistent with prior estimates; earlier predictions should not change. However, in the context of explicit FE calculations, the NN architecture only "sees" the portion of the deformation that has already occurred, not the full history up until localization.

To mimic this behavior, we selected three loading paths from the original dataset and truncated them by removing a portion of the time history at the end. These truncated paths were then used for predictions with the encoder-decoder-based RNN architectures. We observed that the NN’s response was different when using the truncated history compared to the full history, as shown in Fig. \ref{fig:Truncation} (A). This is not compatible with the physical state of the deformation history, since the damage evolves at each time step in the simulation, and the damage output at each step is unique, remaining unchanged based on future inputs.

On the other hand, when we check the results from the transformer-based NN architecture, we do not observe any changes in the prediction outputs, as shown in Fig. \ref{fig:Truncation} (B). This behavior aligns more closely with the physical state of the deformation history, where earlier damage estimates remain consistent even as the history evolves.

Therefore, while the encoder-decoder-based RNNs exhibited the best performance in learning deformation history, they cannot be used as a surrogate due to the incompatibility of their mathematical process with the simulation data. This issue arises from the nature of the encoder-decoder architecture: the second layer uses the encoded vector calculated from the input. When the network receives truncated history inputs, it generates different encoded vectors, which ultimately lead to variations in the output predictions.

\section{Conclusions}

In this study, we applied and trained three different sequential learning structures on bilinear loading paths to assess their learning capabilities and suitability as surrogates in FEM simulations. While other surrogate models have been introduced in the literature, to the best of our knowledge, this is the first study to compare the applicability of three well-known sequential learning structures and evaluate how their internal mathematical processes align with the physical state of deformation paths. The results demonstrate that while encoder-decoder-based RNNs achieve the best accuracy in learning deformation history, they cannot be used as surrogates because they conflict with the physical state of the deformation paths, leading to incorrect damage predictions. This study contributes to the future development of surrogate models for materials science and engineering problems and lays the groundwork for future advancements in this field.

\subsubsection{Acknowledgements} This work has  been financially supported by the Estonian Research Council via grant PSG754 (Coupled simulation model for ship crashworthiness assessment). These funding mechanisms are gratefully acknowledged.

\subsubsection{Data availability}{The dataset used in this study is available via link:  

\url{https://doi.org/10.5281/zenodo.14875977}.}

\bibliographystyle{splncs04}
\bibliography{bib}
%Author, F.: Article title. Journal \textbf{2}(5), 99--110 (2016)

%Author, F., Author, S.: Title of a proceedings paper. In: Editor,
%F., Editor, S. (eds.) CONFERENCE 2016, LNCS, vol. 9999, pp. 1--13.
%Springer, Heidelberg (2016). \doi{10.10007/1234567890}

%Author, F., Author, S., Author, T.: Book title. 2nd edn. Publisher,

%Author, A.-B.: Contribution title. In: 9th International Proceedings
%on Proceedings, pp. 1--2. Publisher, Location (2010)

%LNCS Homepage, \url{http://www.springer.com/lncs}. Last accessed 4
%\end{thebibliography}
\end{document}